\documentclass[10pt,twocolumn,letterpaper]{article}

\usepackage[pagenumbers]{cvpr} 

\definecolor{cvprblue}{rgb}{0.21,0.49,0.74}
\usepackage[pagebackref,breaklinks,colorlinks,allcolors=cvprblue]{hyperref}
\usepackage{algorithm}
\usepackage{algpseudocode}
\usepackage{multirow}
\definecolor{LightCyan}{rgb}{0.88,1,1}
\usepackage{color, colortbl}

\def\paperID{*****} 
\def\confName{CVPR}
\def\confYear{2026}

\title{RTGen: Real-Time Generative Detection Transformer}

\author{
Chi Ruan$^{1}$ \qquad
Jiying Zhao$^{2}$ \qquad
Wenhu Chen$^{1}$ \\ [0.5em]
$^1$University of Waterloo \qquad
$^2$University of Ottawa \\
{\tt\small cruan059@uottawa.ca, jzhao@uottawa.ca, wenhuchen@uwaterloo.ca}
}

\begin{document}
\maketitle
\begin{abstract}

Although open-vocabulary object detectors can generalize to unseen categories, they still rely on predefined textual prompts or classifier heads during inference. Recent generative object detectors address this limitation by coupling an autoregressive language model with a detector backbone, enabling direct category name generation for each detected object. However, this straightforward design introduces structural redundancy and substantial latency. In this paper, we propose a \textbf{R}eal-\textbf{T}ime \textbf{GEN}erative Detection Transformer (RTGen), a real-time generative object detector with a succinct encoder-decoder architecture. Specifically, we introduce a novel Region–Language Decoder (RL-Decoder) that jointly decodes visual and textual representations within a unified framework. The textual side is organized as a Directed Acyclic Graph (DAG), enabling non-autoregressive category naming. Benefiting from these designs, RTGen-R34 achieves 131.3 FPS on T4 GPUs, over 270× faster than GenerateU. Moreover, our models learn to generate category names directly from detection labels, without relying on external supervision such as CLIP or pretrained language models, achieving efficient and flexible open-ended detection.
\end{abstract}    
\section{Introduction}
\label{sec:intro}
Object detection \cite{ren2016faster, lin2017feature, redmon2016you, carion2020end} has traditionally been limited to a closed set of predefined categories, which motivated the development of open-vocabulary detection (OVD) techniques. Despite remarkable advances \cite{li2022grounded, gu2021open, zhou2022detecting, minderer2022simple}, existing OVD models still depend on a limited set of predefined textual prompts during inference, constraining their scalability and adaptability in real-world scenarios. To address this limitation, generative object detection \cite{yao2024detclipv3, lin2024generative} has recently emerged as a promising paradigm that enables open-ended category generation—allowing detectors to produce free-form textual descriptions beyond fixed categories.

Several approaches \cite{wu2025grit, lin2024generative, yao2024detclipv3, johnson2016densecap} have explored this direction, though often employing relatively simple designs. For example, GenerateU \cite{lin2024generative} feeds object queries from Deformable DETR \cite{zhu2020deformable} into an autoregressive language model to generate object names, while DetCLIPv3 \cite{yao2024detclipv3} attaches an object captioner to an open-vocabulary detector to produce region-level captions. In general, these methods transmit object features extracted by a detector into a separate language model to generate textual outputs. Although effective, such decoupled architectures exhibit weak integration between detection and language generation, leading to redundant computation and limited cross-modal synergy.

\begin{figure}[t]
\centering
\includegraphics[width=1\linewidth]{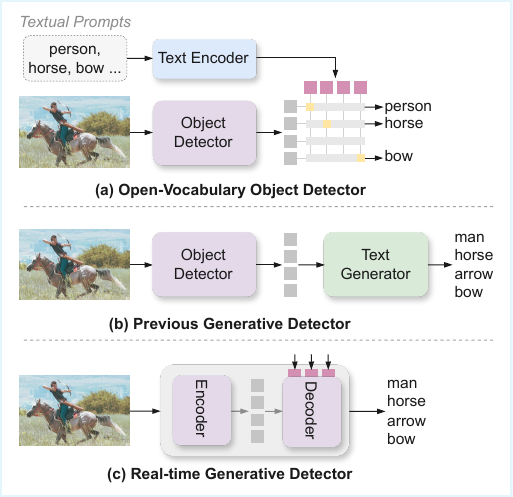}
   \caption{Comparison of different object detection paradigms. The open-vocabulary detector (a) relies on predefined textual prompts and text encoding; previous generative detectors (b) attach an autoregressive text generator to the detector; our real-time generative detector (c) unifies detection and text generation within a single framework, generating category names efficiently without predefined prompts or external language models.}
\label{fig: comparison}
\end{figure}

Inspired by extensive multimodal research \cite{li2023blip, bao2022vlmo, li2022blip, lu2019vilbert} demonstrating the effectiveness of shared transformer architectures for unified vision–language understanding, we propose a Region–Language Decoder (RL-Decoder) that integrates object detection and text generation within a single decoding framework. Unlike previous generative detectors that attach an external language model, our design treats object queries and text embeddings as co-evolving representations processed jointly through cross-modal attention. This unified formulation allows the model to dynamically align visual and textual features at each decoding layer, thereby enhancing cross-modal interaction and eliminating redundant computation between separate modules.

Nevertheless, existing language modeling paradigms are predominantly autoregressive \cite{radford2019language, touvron2023llama, chowdhery2023palm}, which rely on sequential decoding and ground-truth supervision—both incompatible with concurrent reasoning and real-time inference. To address this, we reformulate text generation within the RL-Decoder as a non-autoregressive Directed Acyclic Graph (DAG) generation process, inspired by \cite{gu2017non}. By structuring text embeddings as a DAG to capture token dependencies in parallel, our approach enables simultaneous reasoning over objects and textual concepts, achieving efficient generation without requiring pre-aligned textual inputs.

Building upon the unified RL-Decoder and the DAG-based non-autoregressive generation mechanism, we introduce the Real-Time GENerative Detection Transformer (RTGen), a unified and efficient framework for open-ended object detection. As illustrated in Fig.~\ref{fig: comparison}, unlike open-vocabulary detectors \cite{cheng2024yolo, li2022grounded, liu2025grounding, zhang2022glipv2} that rely on predefined category prompts or previous generative detectors \cite{wu2025grit, lin2024generative, yao2024detclipv3, long2023capdet} that generate text only after detection, RTGen performs joint detection and text generation within a single decoding process. Built upon the efficient RT-DETR architecture \cite{zhao2024detrs}, our model incorporates the proposed Region–Language Decoder (RL-Decoder) and a non-autoregressive DAG-based generation mechanism, enabling concurrent reasoning over visual and textual representations for efficient and open-ended detection.

In summary, our main contributions are as follows:
\begin{itemize}
\itemsep0em
\item We propose the Real-Time GENerative Detection Transformer (RTGen), a unified and efficient framework that achieves real-time open-ended detection without relying on predefined categories, making it suitable for practical deployment.
\item We design a novel Region–Language Decoder (RL-Decoder) that enables concurrent reasoning over visual and textual representations within a single framework, and further produces category names in a non-autoregressive manner for efficient open-ended generation.
\item Through these designs, RTGen-R34 reaches 131.3 FPS on T4 GPUs with TensorRT FP16, compared with 0.48 FPS on GenerateU, and RTGen-R101 achieves 40.7 AP on COCO, without relying on CLIP or any pretrained language models.
\end{itemize}

\section{Related Works}
\label{sec:related}

\subsection{Open-Vocabulary Object Detection} 

Open-vocabulary object detection (OVD), which aims to recognize objects beyond a fixed set of predefined categories, has recently received increasing attention. Most OVD methods \cite{li2022grounded, minderer2022simple, cheng2024yolo, liu2025grounding} integrate a pre-trained language encoder into the detector and classify regions by measuring visual–textual similarity. A parallel line of work \cite{zareian2021open, zhong2022regionclip, gu2021open} distills semantic knowledge from vision–language models (VLMs) to enhance the detector’s open-set recognition ability. For example, OVR-CNN \cite{zareian2021open} is an early attempt that leverages grounded image–caption pre-training to align visual and textual representations, while ViLD \cite{gu2021open} transfers knowledge from the CLIP model \cite{radford2021learning} to a two-stage detector via feature-level distillation. Another strategy for improving OVD performance is to expand the training data through vision–language pre-training. GLIP \cite{li2022grounded} unifies object detection and phrase grounding into a single pre-training framework, and Grounding DINO \cite{liu2025grounding} further integrates grounded pre-training into a detection transformer, using multimodal fusion modules to strengthen visual–textual interaction.

Despite significant progress, existing OVD models still require human-provided category names at inference time. This dependence on predefined prompts fundamentally restricts their scalability, adaptability, and applicability in truly open-world scenarios.

\subsection{Dense Captioning} 

Dense captioning aims to generate detailed descriptions for specific areas of an image. The task was introduced by Johnson et al. through FCLN \cite{johnson2016densecap}, a fully convolutional localization network that extracts region proposals with a CNN and localization layer, processes them with a recognition network, and decodes captions using an RNN. CapDet \cite{joseph2021towards} extends this idea to open-world scenarios, feeding unlabeled object proposals into a captioning head to produce free-form descriptions. More recent approaches, such as GRiT \cite{wu2025grit} and DetCLIPv3 \cite{yao2024detclipv3}, attach a text generator to an object detector to produce object descriptions. While dense captioning focuses on generating rich, descriptive captions for objects or regions, our setting instead emphasizes predicting concise, canonical category names. Dense captioning is typically evaluated using both AP from object detection and METEOR \cite{banerjee2005meteor} from machine translation.

\begin{figure*}[!t]
\centering
\includegraphics[width=1\linewidth]{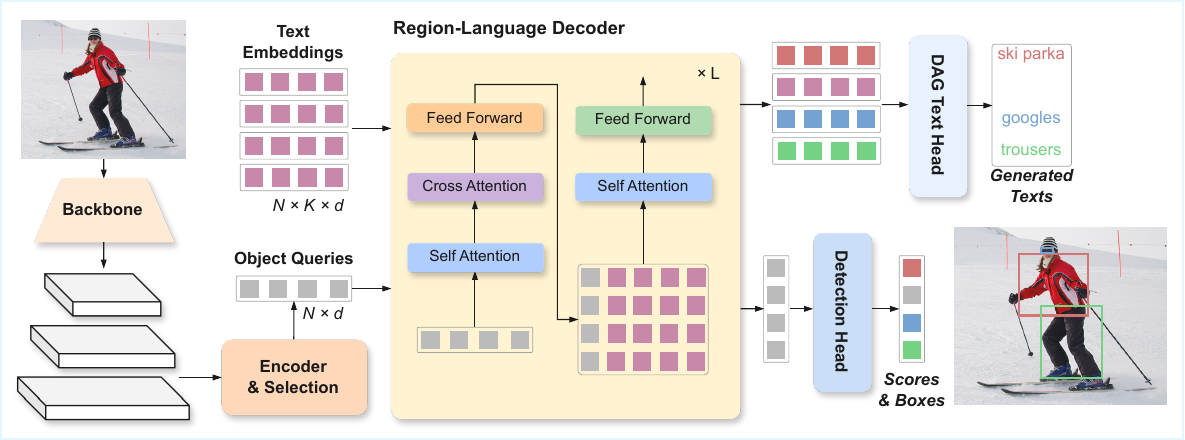}
   \caption{Overall architecture of RTGen. RTGen builds upon RT-DETR by introducing a unified Region–Language Decoder (RL-Decoder) that jointly processes object queries and positional text embeddings. The refined queries and text features are sent to a detection head and the DAG Text Head, enabling efficient real-time open-ended detection.}
\label{fig: network}
\end{figure*}

\begin{figure}[!t]
\centering
\includegraphics[width=1\linewidth]{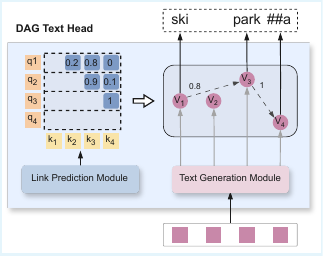}
   \caption{Structure of the proposed DAG Text Head. It estimates token transition probabilities via the Link Prediction Module and constructs a directed acyclic graph for non-autoregressive text generation.}
\label{fig: text head}
\end{figure}

\subsection{Non-autoregressive Translation} 

Non-autoregressive translation (NAT) models aim to generate sequences in parallel without conditioning on previously produced tokens, greatly accelerating inference. The idea was introduced by Gu et al. \cite{gu2017non}, though the removal of sequential dependencies leads to a clear accuracy gap relative to autoregressive (AT) models. Subsequent approaches seek to balance speed and dependency modeling: SAT \cite{wang2018semi} adds limited autoregressive steps, CMLM \cite{ghazvininejad2019mask} iteratively refines masked tokens, and DA-Transformer \cite{huang2022directed} uses a directed acyclic graph (DAG) to organize token generation in a fully NAT manner, enabling parallel decoding while preserving structural dependencies. Inspired by this DAG formulation, RTGen adopts a lightweight DAG-based text head that enables efficient non-autoregressive category-name generation, preserving both consistency and dependency awareness among output tokens.
\section{Method}
\label{sec:method}

In this section, we present RTGen, a real-time generative object detection transformer. To achieve efficient integration of detection and language generation, we propose a Region–Language Decoder (RL-Decoder) that concurrently processes object queries and textual representations within a unified decoding structure. Furthermore, the decoder incorporates a DAG-based non-autoregressive text generation mechanism, inspired by \cite{huang2022directed}, enabling parallel generation of category names alongside object decoding. The overall architecture of RTGen is illustrated in Fig.~\ref{fig: network}. We describe the model architecture in Sec.~\ref{Model Architecture}, the RL-Decoder in Sec.~\ref{RL-Decoder}, the DAG Text Head in Sec.~\ref{DAG}, and the training formulation in Sec.~\ref{Training Formulation}.

\subsection{Model Architecture} \label{Model Architecture}

We adopt RT-DETR~\cite{zhao2024detrs}, a closed-set object detector, as the foundation of our architecture. As a transformer-based detector derived from DETR~\cite{carion2020end}, it employs an encoder–decoder design rather than a conventional CNN backbone, offering stronger compatibility for integration with language models. Moreover, RT-DETR achieves an excellent balance between accuracy and efficiency, outperforming both traditional DETR and Deformable-DETR~\cite{zhu2020deformable} while maintaining real-time performance. It incorporates an efficient hybrid encoder that disentangles intra-scale interactions and cross-scale integration among different features, as well as a decoder with deformable attention~\cite{zhu2020deformable} for content-aware sampling. In this work, we disable its classification functionality and adopt a class-agnostic detection setting, focusing solely on object detection and localization.

\begin{algorithm}[t]
\caption{Region-Language Decoder Forward Process}
\label{alg:decoder}
\textbf{Input:} object queries $\mathbf{Q} \in \mathbb{R}^{N \times d}$, 
image features $\mathbf{M} \in \mathbb{R}^{M \times d}$, 
language embeddings $\mathbf{T} \in \mathbb{R}^{N \times K \times d}$, 
number of layers $L$ \\
\textbf{Output:} Updated queries $\mathbf{Q}_{L+1}$ and text features $\mathbf{T}_{L+1}$

\begin{algorithmic}[1]
\State \textbf{Initialize:} $\mathbf{Q}_1 \gets \mathbf{Q}$, $\mathbf{T}_1 \gets \mathbf{T}$
\For{$l = 1$ to $L$}

    \State \textbf{Region-aware update:}
    \begin{equation}\label{eq:region}
        \mathbf{Q}_{l+1} = \text{FFN}_{r}\!\left(\text{CrossAttn}\!\left(\text{SelfAttn}\!\left(\mathbf{Q}_{l}\right), \mathbf{M}\right)\right)
    \end{equation}

    \State \textbf{Language fusion:}
    \begin{equation}\label{eq:concat}
        \mathbf{H}_{l} = \text{Concat}_{\text{dim}=2}\!\left(\mathbf{Q}_{l+1}, \mathbf{T}_{l}\right)^{\top_{(1,2)}}
    \end{equation}

    \State \textbf{Cross-modal interaction:}
    \begin{equation}\label{eq:cross-modal}
        \tilde{\mathbf{H}}_l = \text{FFN}_{c}(\text{SelfAttn}(\mathbf{H}_l))
    \end{equation}

    \State \textbf{Feature separation:}
    \begin{equation}\label{eq:split}
        \_, \mathbf{T}_{l+1} = 
        \text{Split}\!\left(
            \tilde{\mathbf{H}}_{l}^{\top_{(1,2)}}, 
            \text{dim}{=}2, 
            \text{size}{=}[1, K]
        \right)
    \end{equation}

\EndFor
\end{algorithmic}
\end{algorithm}

The overall architecture of our proposed RTGen is illustrated in Fig.~\ref{fig: network}. Built upon RT-DETR~\cite{zhao2024detrs}, RTGen inherits its real-time efficiency from the retained backbone and encoder, while introducing a novel decoding structure and a text generation head to enable open-ended object detection. The architecture consists of a backbone, an encoder, a Region–Language Decoder (RL-Decoder), a DAG Text Head, and a detection head. The backbone extracts hierarchical visual representations from input images, and the encoder transforms them into context-aware object queries. These object queries are jointly decoded with positional text embeddings in the RL-Decoder, which concurrently enhances object semantics and builds corresponding textual representations, achieving unified reasoning across visual and textual modalities. The updated object queries are sent to the detection head to predict objectness scores and bounding boxes, whereas the refined text embeddings are passed to the DAG Text Head, which organizes them into a directed acyclic graph, models token dependencies via attention, and generates category names in a non-autoregressive manner. Through this integrated architecture, RTGen achieves efficient, real-time open-ended detection with unified object understanding and language generation.

\subsection{RL-Decoder}\label{RL-Decoder}

The decoder serves as the core component for cross-modal fusion, where object queries propagate visual semantics into the corresponding text embeddings. As detailed in Algorithm~\ref{alg:decoder}, each decoder layer refines object queries and fuses them with text embeddings through four sequential operations. First, a region-aware update is performed, where the object queries $\mathbf{Q}_l$ are successively processed by self-attention, cross-attention with image features $\mathbf{M}$, and a feed-forward network $\text{FFN}_{r}$, as formulated in Eq.~\ref{eq:region}. Next, the refined queries $\mathbf{Q}_{l+1}$ are concatenated with the text embeddings $\mathbf{T}_l$ to form a joint representation $\mathbf{H}_l$, as shown in Eq.~\ref{eq:concat}. Cross-modal interaction is then applied via self-attention and another feed-forward network $\text{FFN}_{c}$ to propagate region-level semantics into the text embeddings, as described in Eq.~\ref{eq:cross-modal}. Finally, the fused representation $\tilde{\mathbf{H}}_l$ is transposed and split along the token dimension to obtain the updated text embeddings $\mathbf{T}_{l+1}$, as shown in Eq.~\ref{eq:split}.

Notably, the same self-attention module is shared across both the region-aware and cross-modal stages to maintain consistent contextual reasoning across modalities, while $\text{FFN}_{r}$ and $\text{FFN}_{c}$ are separately parameterized to serve distinct purposes—region-level feature refinement and cross-modal fusion, respectively. Only the queries updated from the region-aware step are propagated to the next layer, preventing textual features from interfering with object queries. This iterative decoding process enables the RL-Decoder to achieve efficient and unified reasoning across visual and language modalities.

\subsection{DAG Text Head}\label{DAG}

After aggregating information from the object queries in the decoder, the text embeddings are passed into the DAG Text Head, which adopts a DAG-based non-autoregressive text generation approach inspired by \cite{huang2022directed}, enabling parallel category name generation. Specifically, the output text embeddings from the decoder, denoted as $\mathbf{T}_{L+1}$, are decomposed into $N$ components $\{\mathbf{T}_{L+1}^{(n)}\}_{n=1}^{N}$, as shown in Eq.~\ref{eq:tensor_decomposition}, where $L$ indicates the layer index of the decoder. Each component $\mathbf{T}_{L+1}^{(n)} \in \mathbb{R}^{K \times d}$ represents a sequence of $K$ token embeddings associated with the $n$-th detected region. For each $\mathbf{T}_{L+1}^{(n)}$, query and key representations are obtained through linear projections using learnable matrices $\mathbf{W}_\text{Q}$ and $\mathbf{W}_\text{K}$, respectively, as defined in Eq.~\ref{eq:qk_definition}. The transition probability matrix $\mathbf{E}$ of the directed acyclic graph is then computed by applying the scaled dot-product attention mechanism, as formalized in Eq.~\ref{eq:attention_matrix}. This matrix captures the directed dependencies among tokens, allowing the DAG Text Head to efficiently model non-autoregressive text generation across all region-specific text embeddings in parallel.

\begin{equation}
    \mathbf{T}_{L+1} \in \mathbb{R}^{N \times K \times d} 
    \quad \Rightarrow \quad 
    \Big\{\, \mathbf{T}_{L+1}^{(n)} \in \mathbb{R}^{K \times d} \,\Big\}_{n=1}^N
    \label{eq:tensor_decomposition}
\end{equation}

\begin{equation}
\mathbf{Q} = \mathbf{T}_{L+1}^{(n)} \mathbf{W}_\text{Q}, \quad 
\mathbf{K} = \mathbf{T}_{L+1}^{(n)} \mathbf{W}_\text{K}.
\label{eq:qk_definition}
\end{equation}

\begin{equation}
\mathbf{E} = \text{softmax}\!\left( \frac{\mathbf{Q} \mathbf{K}^\top}{\sqrt{d}} \right).
\label{eq:attention_matrix}
\end{equation}

A key advantage of this DAG formulation is that it eliminates the need for autoregressive teacher-forcing during training. Instead of conditioning on ground-truth token sequences, the DAG Text Head learns category-level text representations directly from the matched ground-truth annotations during detection training. Specifically, object queries are first aligned with ground-truth boxes using Hungarian matching, and the corresponding category labels serve as supervision signals for learning token dependencies through the attention-based transition matrix. This formulation enables efficient parallel text generation while remaining consistent with the detection training pipeline.

Among the available decoding options for the DAG Text Head, Viterbi decoding~\cite{shao2022viterbi} offers a principled and reliable way to select the final path by performing dynamic programming over the predicted token graph. Unlike greedy decoding, which commits to locally optimal transitions, or lookahead decoding, which only partially anticipates future likelihoods, Viterbi jointly evaluates all transitions to recover a globally coherent sequence. Compared with sampling-based approaches, it avoids randomness and yields deterministic outputs, while also being more efficient than beam search, which requires maintaining multiple parallel hypotheses. Given these advantages in stability and efficiency, all subsequent experiments in this paper use Viterbi decoding as the default strategy.

\subsection{Training Formulation}
\label{Training Formulation}

\textbf{Training Data.}
In traditional object detection, the training sample is defined as \( (x, \{\mathbf{b}_i, c_i\}_{i=1}^{N}) \), where \( x \in \mathbb{R}^{3 \times H \times W} \) is the input image, \( \{\mathbf{b}_i|\mathbf{b}_i \in \mathbb{R}^4\}_{i=1}^N \) denotes the bounding boxes, and \( \{c_i\}_{i=1}^N \) are discrete category labels. In contrast, we reformulate the data as \( (x, \{\mathbf{b}_i, y_i\}_{i=1}^{N}) \), where \( \{y_i\}_{i=1}^N \) denotes the category name in text form. Unlike open-vocabulary detectors that require predefined category names as input queries, RTGen directly takes the image \( x \) and learns to predict both boxes \( \{\hat{\mathbf{b}}_j\}_{j=1}^{K} \) and the corresponding generated texts \( \{\hat{y}_j\}_{j=1}^{K} \).

\textbf{Training Objective.}
Given the supervision \( \{\mathbf{b}_i, y_i\} \), RTGen jointly optimizes localization and text generation through a unified objective. For each matched prediction \( (\hat{\mathbf{b}}_j, \hat{y}_j) \), we apply a box regression loss \( \mathcal{L}_{\text{reg}} \), an IoU-based localization loss \( \mathcal{L}_{\text{iou}} \), a DAG text generation loss \( \mathcal{L}_{\text{DAG}} \), and an objectness loss \( \mathcal{L}_{\text{obj}} \). The Hungarian matching cost uses the predicted objectness score instead of a classification score, consistent with our class-agnostic setting. Furthermore, \( \mathcal{L}_{\text{DAG}} \) is dynamically scaled by the IoU between \( \hat{\mathbf{b}}_j \) and its matched ground-truth \( \mathbf{b}_i \), amplifying high-quality matches while reducing the influence of poorly localized predictions.

\section{Experiments}
\label{sec:experiments}

\begin{table*}[]
\centering
\caption{Comparison of closed-set, open-set, and open-ended object detection methods on the COCO validation set \cite{lin2014microsoft}. RTGen achieves competitive accuracy while significantly outperforming prior open-ended models in speed. The FPS is measured on an NVIDIA T4 GPU using TensorRT FP16 optimization, with an input size of (640, 640). Dataset abbreviations: OI denotes OpenImages \cite{krasin2017openimages}, O365 denotes Objects365 V1 \cite{shao2019objects365}, and VG denotes Visual Genome \cite{krishna2017visual}.}
\label{tab:main_comparison}
\resizebox{\textwidth}{!}{
\begin{tabular}{ccccccccc}
\toprule
Method       & BackBone & Type  & Supervision     & Training Data   & \#Params (M) & GFLOPs & FPS    & AP   \\
\midrule
Faster R-CNN \cite{girshick2015fast} & R50      & Closed-Set & -  & COCO            & 42           & 180    & -      & 40.2 \\
Faster R-CNN \cite{girshick2015fast} & R101     & Closed-Set & -   & COCO            & 60           & 246    & -      & 42   \\
DETR-DC5 \cite{carion2020end}    & RN50     & Closed-Set & -   & COCO            & 41           & 187    & -      & 43.3 \\
DETR-DC5 \cite{carion2020end}    & R101     & Closed-Set & -   & COCO            & 60           & 253    & -      & 44.9 \\
\midrule
OWL-ViT \cite{minderer2022simple}     & ViT-B    & Open-Set & CLIP   & OI, VG          & -            & -      & -      & 30.3 \\
OWL-ViT  \cite{minderer2022simple}    & ViT-L    & Open-Set & CLIP  & OI, VG          & -            & -      & -      & 34.7 \\
ViLD  \cite{gu2021open}       & R50      & Open-Set & CLIP  & LVIS base       & -            & -      & -      & 36.6 \\
OV-DETR \cite{zang2022open}     & R50      & Open-Set & CLIP  & LVIS base       & -            & -      & -      & 38.1 \\
\midrule
GenerateU \cite{lin2024generative}   & Swin-T   & Open-Ended & FlanT5-base & VG              & 297          & -      & 0.48   & 33.0 \\
GenerateU  \cite{lin2024generative}  & Swin-T   & Open-Ended & FlanT5-base & VG, GRIT        & 297          & -      & 0.48   & \textbf{33.6} \\
RTGen        & R50      & Open-Ended & - & O365      & 71           & 225    & \textbf{90.4}  & 32.0 \\
RTGen        & R101     & Open-Ended & - & O365      & 105          & 348    & 59.7  & \textbf{33.6} \\
\midrule
RTGen        & R50      & Open-Ended & - & O365+COCO & 71           & 225    & \textbf{90.4}  & 39.7 \\
   RTGen     & R101     & Open-Ended & - & O365+COCO & 105          & 348    & 59.7  & \textbf{40.7} \\
\bottomrule
\end{tabular}}
\end{table*}

In this section, we present a comprehensive evaluation of the proposed RTGen. We first describe the experimental settings, including datasets, the evaluation protocol, and implementation details. Subsequently, we report the performance of RTGen under various metrics and present ablation studies to examine the impact of each component in our framework.

\subsection{Experimental Settings}
\textbf{Datasets.} We conduct experiments on two large-scale detection datasets: COCO \cite{lin2014microsoft} and Objects365 V1 \cite{shao2019objects365}. COCO contains 80 object categories with diverse scenes and dense annotations, while Objects365 provides 365 categories with significantly larger coverage and richer contextual diversity. These complementary datasets allow us to comprehensively evaluate RTGen, where we only utilize their category names without using the original classification labels.

\textbf{Evaluation Protocol.} Evaluating open-ended category generation is challenging, as a single object may correspond to multiple semantically valid names, such as `couch' and `sofa'. Direct string matching cannot capture such semantic flexibility. Following \cite{lin2024generative}, we measure similarity in a continuous embedding space instead of relying on exact lexical matches. A pretrained text encoder projects both generated names and ground-truth annotations into a shared semantic space; each generated name is matched to the ground-truth category with the highest similarity, and this maximum similarity is used to rescale the model’s prediction score. The text encoder is used only for evaluation. In our experiments, we adopt the CLIP text encoder \cite{radford2021learning} for its stable and widely used embedding space.

\textbf{Implementation Details.} Following the settings of RT-DETR \cite{zhao2024detrs}, we train three variants of RTGen with
ResNet-34, ResNet-50, and ResNet-101 \cite{he2016deep} backbones, where the hidden dimensions are set to
256, 256, 384, respectively. The decoder consists of
6 layers with 8 attention heads. We select
300 object queries from the encoder features, and each query is
associated with 8 text embeddings for DAG-based generation. We use AdamW \cite{loshchilov2017decoupled} as the optimizer with a base learning rate
\(1\times10^{-4}\), a backbone learning rate
\(1\times10^{-5}\), \(\beta\) values of 0.9 and 0.999, and a weight decay of \(1\times10^{-4}\).
All models are trained for 72 epochs on COCO and 12 epochs on Objects365, using a batch size of 16.
The overall training objective is formulated as
\[
\mathcal{L}
= \lambda_{\text{reg}} \mathcal{L}_{\text{reg}}
+ \lambda_{\text{iou}} \mathcal{L}_{\text{iou}}
+ \lambda_{\text{obj}} \mathcal{L}_{\text{obj}}
+ \lambda_{\text{DAG}} \mathcal{L}_{\text{DAG}},
\]
where the loss weights are set to
\(\lambda_{\text{reg}} = 5.0\),
\(\lambda_{\text{iou}} = 2.0\),
\(\lambda_{\text{obj}} = 1.0\),
and \(\lambda_{\text{DAG}} = 1.0\).
The DAG loss is computed without normalizing by the number of ground-truth boxes.

\subsection{Generative Object Detection}

Table \ref{tab:main_comparison} compares RTGen with representative closed-set, open-set, and open-ended detectors on COCO. Unlike existing open-set and open-ended approaches that rely on massive external semantic supervision, such as CLIP \cite{radford2021learning}, trained on hundreds of millions of image–text pairs, or large language models like FlanT5-base \cite{chung2024scaling}, RTGen does not use any external text or vision–language pre-training. Instead, it learns to produce category names directly from the annotations provided by standard detection datasets, and this generative capability emerges even though the matched training boxes are not perfectly aligned with ground-truth boxes. 

With 71 million to 105 million parameters and a computational cost between 225 and 348 GFLOPs, RTGen is substantially lighter than GenerateU, which contains nearly 300 million parameters and requires significantly more computation. This compact design provides a strong efficiency advantage: RTGen-R50 reaches 90.4 FPS on a single T4 GPU using TensorRT FP16, more than 180× faster than GenerateU, while RTGen-R101 also maintains real-time performance. In the zero-shot setting, where the model is trained only on Objects365 and evaluated on COCO, RTGen achieves 32.0 AP for the R50 backbone and 33.6 AP for the R101 backbone, already comparable to prior open-ended detectors such as GenerateU, whose accuracy ranges from 33.0 to 33.6 AP. When trained with the combined Objects365 and COCO datasets, RTGen further improves to 39.7 AP for R50 and 40.7 AP for R101. These results demonstrate that strong generative object detection performance can be achieved efficiently using only standard detection annotations.

Table~\ref{tab:coco_variants} summarizes the performance of RTGen variants trained and evaluated on COCO. As the backbone scales from R34 to R101, the models show steady improvements across all metrics, with AP increasing from 34.8 to 38.8 and reaching 39.6, alongside similar gains in AP$_{50}$ and AP$_{75}$. These trends show that larger backbones provide stronger visual features, which in turn lead to better detection quality and category-name generation. At the same time, all variants maintain high efficiency: even the largest R101 model runs at nearly 60 FPS, while the lightweight R34 model exceeds 130 FPS. Overall, these results show that RTGen scales well with backbone size, improving accuracy while still maintaining real-time speed.

In the zero-shot setting, RTGen demonstrates strong generalization to categories that are never exposed during training. Trained solely on Objects365 and evaluated on COCO novel classes, RTGen surpasses representative open-set approaches, as summarized in Table~\ref{tab: zero_shot_coco}. Specifically, RTGen achieves 35.5 AP{$_{50}$} with an R50 backbone and 37.0 AP{$_{50}$} with R101—substantially higher than prior open-set detectors. These gains are obtained without any external vision–language pre-training, relying only on the category names provided in Objects365. Meanwhile, RTGen maintains competitive performance on COCO base categories and overall metrics. These results highlight the strength of the generative formulation of RTGen in transferring knowledge across datasets and recognizing previously unseen object categories.

\begingroup
\setlength\tabcolsep{10pt}
\begin{table}[t]
\centering
\caption{
Performance of RTGen variants trained and evaluated on the COCO dataset. 
We report inference speed (FPS) together with AP, AP$_{50}$, and AP$_{75}$ scores.
}
\label{tab:coco_variants}
\resizebox{\linewidth}{!}{
\begin{tabular}{lcccc}
\toprule
Method   & FPS & AP & AP$_{50}$ & AP$_{75}$ \\
\midrule
RTGen-R34  & 131.3 & 34.8 & 46.7  & 37.6  \\
RTGen-R50  & 90.4    & 38.8 & 51.2  & 42.0  \\
RTGen-R101 & 59.7 & 39.6 & 52.2  & 43.0  \\
\bottomrule
\end{tabular}}
\end{table}
\endgroup

\begin{table}[]
\centering
\caption{Zero-shot generative object detection results on COCO. Numbers in gray indicate models trained only on COCO’s base categories.}
\label{tab: zero_shot_coco}
\resizebox{\linewidth}{!}{
\begin{tabular}{ccccc}
\toprule
\multirow{2}{*}{Method}    & \multirow{2}{*}{Type}    & Novel & Base & Overall \\
& & AP$_{50}$ & AP$_{50}$ & AP$_{50}$ \\
\bottomrule
OVR-CNN \cite{zareian2021open}     & Open-Set   & 22.8  & {\color{black!50} 46.0} & {\color{black!50} 39.9}    \\
Region-CLIP \cite{zhong2022regionclip} & Open-Set        & 26.8  & {\color{black!50} 54.8} & {\color{black!50} 47.5}    \\
ViLD \cite{gu2021open}       & Open-Set        & 27.6  & {\color{black!50} 59.5} & {\color{black!50} 51.3}    \\
Detic \cite{zhou2022detecting}      & Open-Set        & 27.8  & {\color{black!50} 47.1} & {\color{black!50} 42.0}    \\
OV-DETR \cite{zang2022open}    & Open-Set        & 29.4  & {\color{black!50} 61.0} & {\color{black!50} 52.7}    \\
VLDet \cite{lin2022learning}      & Open-Set        & 32.0  & {\color{black!50} 50.6} & {\color{black!50} 45.8}    \\
\midrule
RTGen-R50   & Open-Ended          & 35.5  & 44.3 & 43.9    \\
RTGen-R101  & Open-Ended          & \textbf{37.0}  & 45.9 & 45.6   \\
\bottomrule
\end{tabular}}
\end{table}

\subsection{Ablation Study}

To better understand the design choices of RTGen, we conduct a series of ablation studies focusing on two key architectural factors: the number of text tokens used for category-name generation and the depth of the decoder. All experiments are performed using the RTGen-R50 model trained and evaluated on the COCO dataset.

\begin{figure}[t]
\centering
\caption{We report AP, AP50, and AP75 on the COCO validation set using RTGen-R50 trained on COCO. The results show that using 8 text tokens achieves the best overall performance across all three metrics.}
\includegraphics[width=1\linewidth]{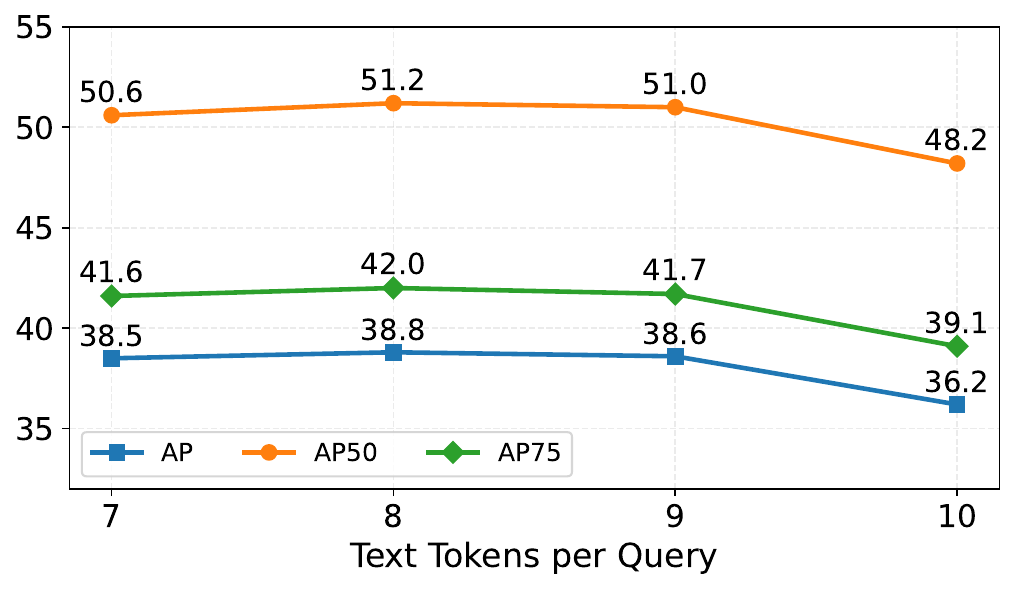}
\label{fig: text_tokens}
\end{figure}

\textbf{Effect of Text Tokens per Query.} To evaluate how the number of text tokens affects generative prediction quality, we vary the number of text tokens from 7 to 10 and report the corresponding AP metrics (AP, AP50, AP75) on COCO. As shown in Fig.~\ref{fig: text_tokens}, the model achieves the best performance when using 8 tokens, reaching 38.8 AP, 51.2 AP50, and 42.0 AP75. Increasing the token length to 9 yields comparable results, suggesting that moderately long category descriptions are beneficial.
However, further expanding the token budget to 10 tokens leads to a noticeable degradation, with AP dropping to 36.2, suggesting that overly long text sequences make it difficult for the object queries in the RL-Decoder to effectively propagate semantic information. These results demonstrate that a compact and well-structured description (8 tokens) provides the best balance between expressiveness and discriminability, enabling RTGen to generate category names that are both informative and tightly grounded to object regions.

\textbf{Effect of Decoder Depth.} While a 6-layer decoder is known to be optimal in the original RT-DETR architecture, the RL-Decoder in RTGen adopts a different structure that simultaneously refines object queries and processes text-embedding features. This dual responsibility suggests that increasing the decoder depth may provide additional capacity for text-conditioned reasoning, motivating an ablation on the number of layers. As summarized in Table~\ref{tab:decoder_layers}, the 6-layer configuration again delivers the strongest accuracy, achieving 38.8 AP. Reducing the depth leads to a modest drop in performance, whereas further increasing it offers no improvement. Instead, deeper decoders introduce notable overhead: GFLOPs rise from 219 to 238, and FPS decreases from 92.5 to 83.9 when expanding from 5 to 8 layers. Overall, these results indicate that six layers strike an effective balance between accuracy and efficiency for jointly refining object queries and integrating text features, and that additional depth yields diminishing returns.

\begin{table}[t]
\centering
\caption{Ablation study on the number of decoder layers.}
\centering
\label{tab:decoder_layers}
\setlength\tabcolsep{6pt}
\begin{tabular}{c|cccc}
\toprule
Layers & AP & Params & GFLOPs & FPS \\
\midrule
5 & 38.2  & 70 & 219 & 92.5 \\
6 & 38.8  & 71 & 225 & 90.4 \\
7 & 38.5  & 73 & 232 & 86.5 \\
8 & 38.2  & 75 & 238 & 83.9 \\
\bottomrule
\end{tabular}
\end{table}

\begin{figure*}[!t] 
    \centering
    \includegraphics[page=1, width=\textwidth]{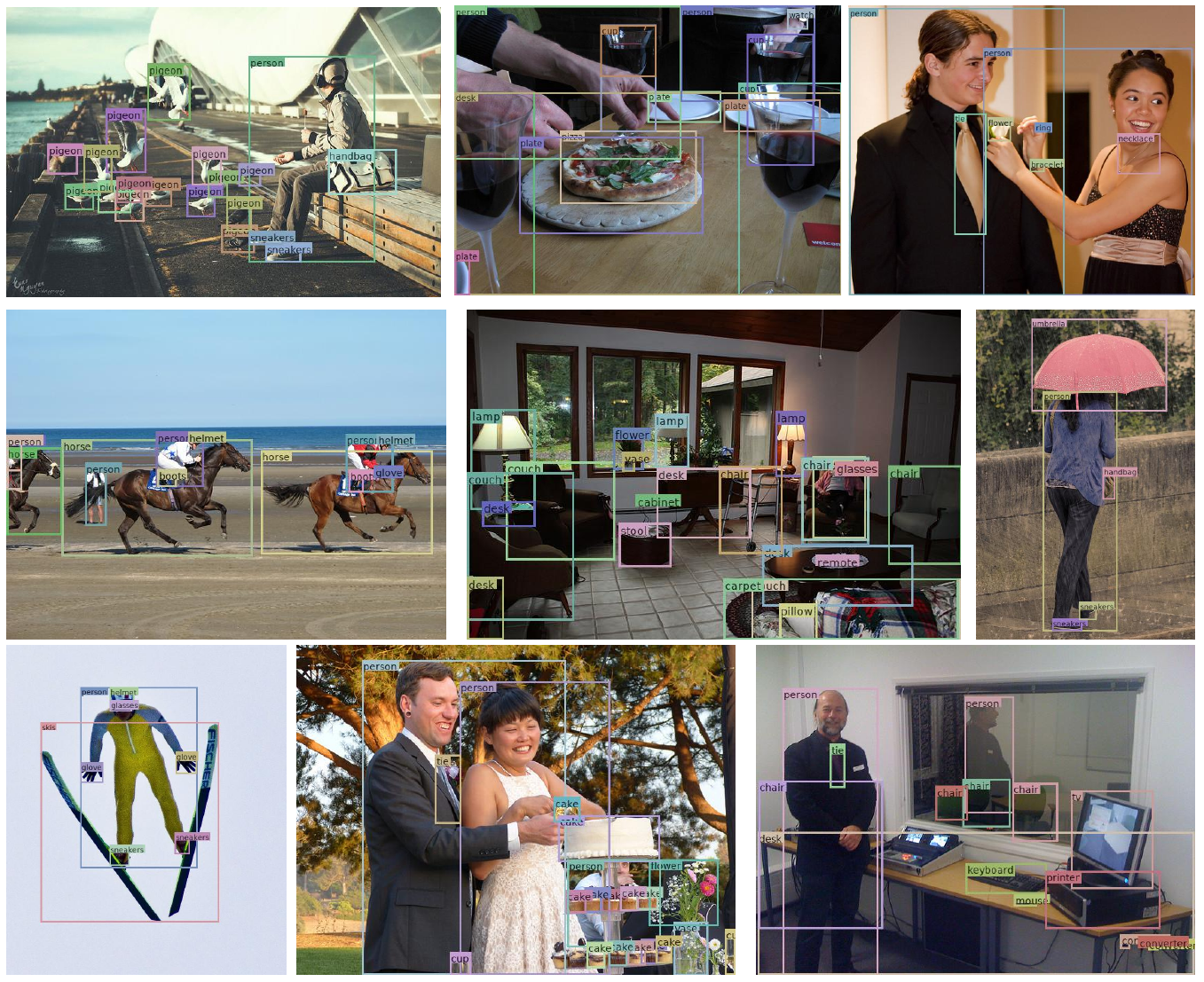}
    \caption{Visualization results from RTGen-R101, trained on Objects365 and inferred in a zero-shot setting on the COCO val.}
    \label{fig: visual}
\end{figure*}

\subsection{Visualization}
In Fig.~\ref{fig: visual}, we present qualitative visualization results produced by RTGen-R101. The model is trained on Objects365 and applied in a zero-shot setting to COCO val images, without using any COCO annotations. These examples illustrate the model’s ability to generalize to unseen concepts and demonstrate consistent generative predictions across diverse scenes.

\subsection{Limitation}
Although RTGen is trained without any external linguistic priors, its vocabulary is inherently limited by the annotations of detection datasets. For example, Objects365 contains only 365 category names, and current detection datasets generally lack large-scale language diversity. As a consequence, the model’s generative capacity is constrained, making it unsuitable for zero-shot evaluation on benchmarks such as LVIS that require a much broader vocabulary, where the dataset vocabulary becomes the dominant bottleneck. Nevertheless, a model trained solely on Objects365 can still generalize reasonably well: when evaluated zero-shot on COCO, RTGen achieves 33.6 AP. This suggests that, given richer linguistic supervision in future detection datasets, RTGen has strong potential to generalize its learned vocabulary to broader domains.
\section{Conclusion}

In this work, we presented RTGen, a real-time generative object detector that integrates category-name generation into real-time detection. Unlike prior open-set or open-ended models that rely on large-scale vision–language pre-training or heavyweight language models, RTGen achieves generative detection using only standard detection annotations. Central to our design is the RL-Decoder, which refines object queries while processing text embeddings, together with a DAG-based generative head that provides fast and consistent category-name decoding. Comprehensive experiments on COCO demonstrate that RTGen delivers competitive accuracy, zero-shot generalization, and substantially higher speed than existing open-ended approaches. Together, these results establish real-time generative detection as a practical and scalable paradigm for open-ended object detection.

{
    \small
    \bibliographystyle{ieeenat_fullname}
    \bibliography{main}
}


\end{document}